# Image matting with normalized weight and semi-supervised learning

*Ping Li, Tingyan Duan, Yongfeng Cao\**
Big data and computer science school
Guizhou Normal University
Guizhou China
1924678362@qq.com, 1181268816@qq.com, cyfeis@whu.edu.cn

*Abstract*—Image matting is an important vision problem. The main stream methods for it combine sampling-based methods and propagation-based methods. In this paper, we deal with the combination with a normalized weighting parameter, which could well control the relative relationship between information from sampling and from propagation. A reasonable value range for this parameter is given based on statistics from the standard benchmark dataset[1]. The matting is further improved by introducing semi-supervised learning iterations, which automatically refine the trimap without user's interaction. This is especially beneficial when the trimap is coarse. The experimental results on standard benchmark dataset have shown that both the normalized weighting parameter and the semi-supervised learning iteration could significantly improve the matting performance.

## I. INTRODUCTION

### A. Background and related work

Matting is an important image processing technology for accurately estimating foreground objects in images and videos. It is often used in image processing software, virtual studio, film post-production and so on[2]. Mathematically, for a given image $I$, any pixel can be expressed as a linear combination of foreground color $F$ and background color $B$:

$$I_z = \alpha_z F + (1-\alpha_z) B, \quad (1)$$

where $z=(x,y)$ represents the pixel coordinates in the image, and $\alpha_z \in [0,1]$ is the foreground opacity of the pixel at z[3]. If $\alpha_z = 1$, the pixel is the foreground; if $\alpha_z = 0$, then the pixel is the background; when $0 < \alpha_z < 1$, the pixel is a mixed pixel, which means the pixel is affected by the foreground pixel and the background pixel at the same time. Usually the most pixels of a natural image are foreground and background, and only a small number of them are mixed pixels[2]. Most matting algorithms need a trimap as input[2]. In a trimap, there are three areas: the foreground area $\Omega_F$ ($\alpha_z=1$), the background area $\Omega_B$ ($\alpha_z=0$) and the unknown area $\Omega_U$ ($0 < \alpha_z < 1$). The main purpose of the matting is to accurately classify the pixels in the unknown area.

Recently, many deep learning methods were used for image matting[4, 5]. Xu etc[4]. first predicted the initial alpha matte with an deep convolutional encoder-decoder neural network, then made a further refinement for the initial alpha matte with an small convolution neural network. Shen etc[5]. Proposed an automatic matting method for portrait using convolution neural network.

Except for above deep-learning-based ones, most of the image matting methods could be categorized into sampling-based methods[3, 6], propagation-based methods[7] and combination of the two methods[8, 9]. Sampling-based methods need to collect sample pairs similar to the unknown pixels from the foreground area and the background area based on pixel color similarity. If the input image has no obvious foreground color and background color or has highly textured regions, this kind of methods are less effective. The latest work [6] makes up for this shortcoming by applying local smoothing as a post-processing step to further improve the quality of alpha matte. Propagation-based methods propagate alpha values of the known pixels to the unknown pixels through local smoothing. They could work on texture images, but still not on images with complex structures.

More effective methods are those combining sampling and propagation. The robust matting method[8] selects high confident samples to estimate alpha value and propagates alpha values of known pixels to unknown pixels with a local window. The local and nonlocal smooth priors method[9] adds nonlocal smoothing information, except for the information from sampling and propagation. These methods have achieved good matting effects. But they did not give a reasonable way to balance the data term from sampling and the local smooth term from propagation, setting only an empirical weight on the data term.

### B. The main contributions of this paper

There are two points accounting for the main contributions of this paper:

- A normalized weight parameter is used to well control the relative role of the data term and the local smooth term in matting, and a reasonable value range for setting the parameter is given based on experiments on the standard data set[1].

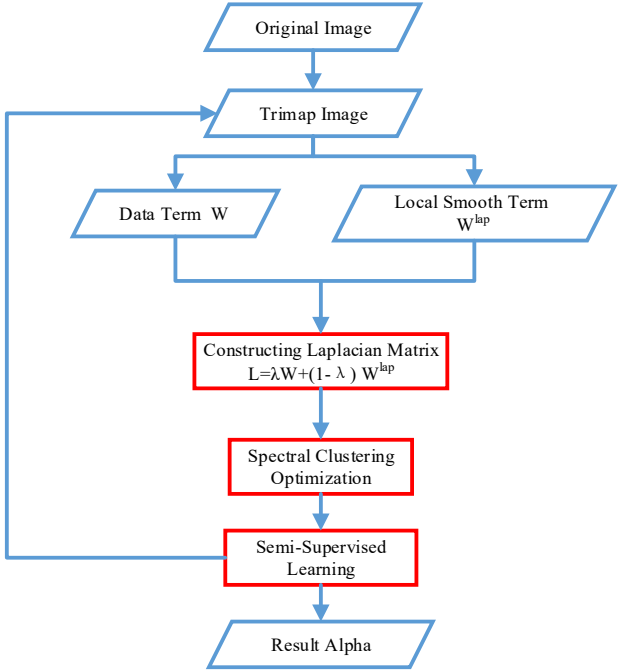

Fig.1. Method flow chart

- Semi-supervised learning iterations are introduced into matting to increase the labeled pixels in the trimap incrementally. It could improve matting effect without increasing the trimap-making burden of users.

*C. Paper content arrangement*

The contents of this paper are arranged as follows: The first section introduces the research background, related works and the main contributions of this paper. The second section is the method part of this paper. In the third section, the method is experimented and analyzed based on the standard data set[1]. The fourth section summarizes this paper and points out the future research direction.

## II. MATTING METHOD

Our method mainly includes three steps as in Fig.1. First, the Laplacian Matrix $L$ is constructed by combining data term $W$ and local smooth term $W^{lap}$ with a normalized weight parameter. Second, alpha matte is got with spectral clustering optimization based on $L$. Third, based on current alpha matte, semi-supervised learning is used to refine the trimap. This makes a loop in the process and enables our method iterating many times to achieve good resulting matte.

*A. Normalized weight parameter*

In order to well control the relatively relationship between data term and local smooth term in the matting process, this paper constructs the Laplacian matrix with a normalized weight parameter $\lambda \in [0,1]$ as follow:

$$L = \lambda W + (1-\lambda)W^{lap}, \quad (2)$$

where $W$ is the data item that is contributed by sampling-based methods, and $W^{lap}$ is the local smooth term that is contributed from propagation-based methods. We calculate these two term as in[8]. In contrast to the Laplacian construction formula $L = \gamma W + W^{lap}$ of the robust matting method[8], it can be seen that our normalized parameter $\lambda$ can more clearly control the relative weight between the two items. The experiment section will suggest a range for setting $\lambda$.

*B. Optimization method*

Image matting can be treated as a graph partition problem which aims at optimally dividing a weighted graph into two or more sub graphs. The spectral clustering method[10] solve this problem as:

$$\arg\min_{q}(\frac{1}{2}q^T L q), \quad (3)$$

where $L$ is the Laplacian matrix constructed in Section A, the vector $q$ records the alpha values of all pixels in the image (where the foreground is 1, the background is 0, and the others to be solved). Rewrite matrix $L$ and vector $q$ as

$$L = \begin{bmatrix} L_k & R \\ R^T & L_u \end{bmatrix}, \quad (4)$$

$$q = \begin{bmatrix} q_k & q_u \end{bmatrix}, \quad (5)$$

where $L_k$ is the Laplacian matrix of the known region (foreground area and background area), $L_u$ is the Laplacian matrix of the unknown region, $q_k$ is the alpha vector of the known region and $q_u$ is the alpha vector of the unknown region. By substituting (4) (5) into (3) and expanding it, it can be seen that $q_u$ can be got by solving following linear optimization problem:

$$L_u q_u = -R^T q_k, \quad (6)$$

in this paper, we use the conjugate gradient method to obtain $q_u$

*C. Semi-supervised learning*

In the trimap given by users, the more detailed the foreground and background be delineated, the better the result of matting is. However, users want to pay as little effort as possible, so we usually obtain very rough trimap, where a large number of pixels are unknown. In this section, we will introduce semi-supervised learning[11] to automatically increase the number of pixels with known labels in the trimap. The specific process is shown in Fig.2. Firstly, based on the current alpha matte, some pixels with high confidence are chosen from the unknown area and labeled automatically. For example, if the alpha value of a selected pixel is close to zero, it will be labeled as the background; if the alpha value is close to one, it will be labeled as the foreground. Then we update the trimap with these newly-labeled pixels and do the matting process again to produce a new alpha matte. Because there are more pixels with known labels in the updated trimap, the resulting alpha matte will be improved this time. To get the maximum i-mprovement, this semi-supervised learning process can be iterated several times. It does not increase workload of users but computers.

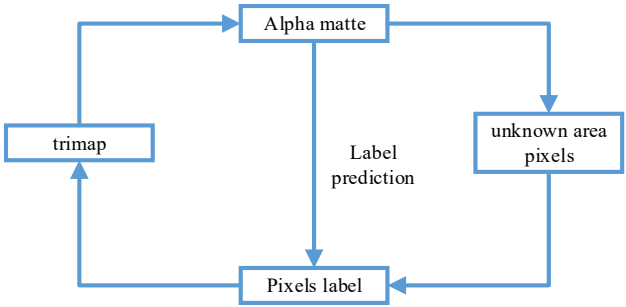

Fig.2. Semi-supervised learning for matting

The wrongly labeled pixels will bring error information into data term and local smooth term, and thus make the alpha matte go wrong. Considering this error, we select pixels from the unknown area with three strict constraints, trying to make sure the predicted labels of these pixels are correct. Namely, for each pixel $x$ that will be updated its label in the trimap by semi-supervised learning, it needs to meet the following conditions at the same time.

**space constraint**:
$$x \in U \text{ and } \exists y \in K, \text{ makes } x \in \phi_y, \quad (7)$$

where $U$ is the set of unknown pixels in current trimap, $K$ is the set of foreground and background pixels in current trimap, and $\phi_y$ is the spatial neighborhood of pixel $y$ (In this paper, we choose the 4-connected pixels).

**confidence constraint**:
$$\alpha_x > t\_\alpha, \quad (8)$$

where $\alpha_x$ is the alpha value of pixel $x$ and $t\_\alpha$ is a threshold.

**proportion constraint**:
$$x \in U_{t\_percent}, \quad (9)$$

where $t\_percent$ is a proportion threshold (In this paper, we use 10%), $U_{t\_percent}$ is the set formed by the top $t\_percent$ pixels of a list that was got by sorting all pixels in current unknown area in descending order according to $|0.5-\alpha_x|$.

## III. EXPERIMENT AND ANALYSIS

The online benchmark for image matting on www.alphamatting.com is used here to evaluate our methods. We choose the low resolution image set which includes 27 input images of size from 490×800 to 719×800, their ground truth and two set of trimaps of different coarse levels. Fig.3 shows sample images in the online benchmark.

The mean square error(MSE) is used as the indicator of m-atting performance in all experiments. The local window used for calculating the local smooth term is set to 3×3.

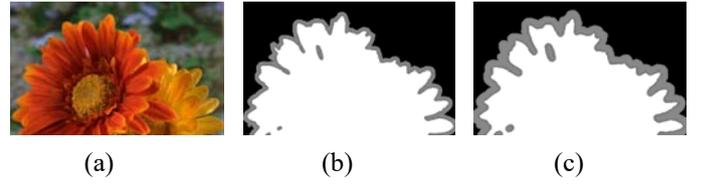

Fig.3 Example images from the online benchmark for image matting. (a) original image; (b) trimap of coarse level 1; (c) trimap of coarse level 2. In the trimaps, the foreground is white, the background is black, and the unknown area is gray.

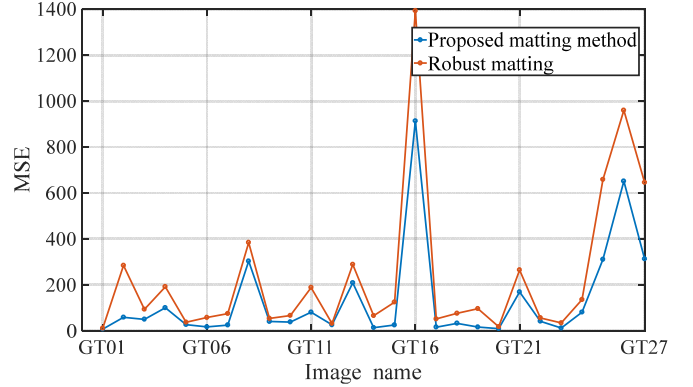

Fig.4. The MSE of proposed matting method ($\lambda$=0.001) and robust matting.

### A. Normalized weight parameter

Fig.4 shows the MSE of our proposed matting method with the normalized parameter $\lambda = 0.001$ and the robust matting method[8] with parameter $\gamma = 0.1$. It can be seen that our method outperforms the robust matting method in almost all test images.

It is important to know how this normalized weight parameter affect matting performance. Fig.5 shows how MSE indicator changes with parameter $\lambda$. It can be seen that MSE indicator is quite stable when parameter $\lambda$ is bigger than 0.05 and all best MSEs for testing images are got with $\lambda$ in [0,0.01]. A rule that the smaller the $\lambda$ is, the better the matting performance is works for all values of $\lambda$ but near zero where the matting performance goes bad when $\lambda$ approaching zero. So we suggest a value range [0.001, 0.01] for setting parameter $\lambda$ in practice.

### B. Semi-supervised learning

In this section, we analyze how semi-supervised learning affect the matting performance and how to choose a good number of iterations. In all experiments, the normalized weight parameter $\lambda$ is set to 0.001. Because semi-supervised learning is especially worth doing when labeled pixels are not enough, for experiments in this section, we use trimaps of coarse level 2(see Fig.3 for the example).

Fig.6 shows the MSE of our proposed matting method without and with semi-supervised learning(iterating 4 times). It can be seen that semi-supervised iterations could improve the matting

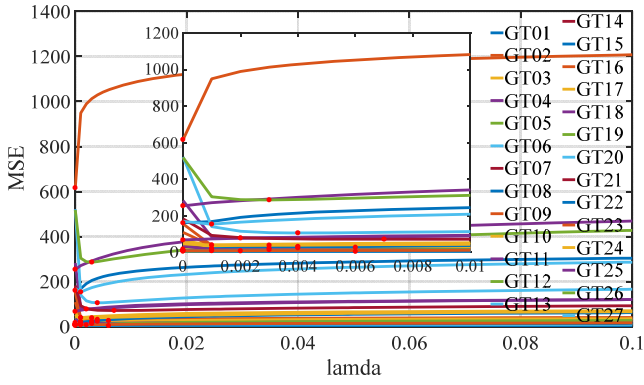

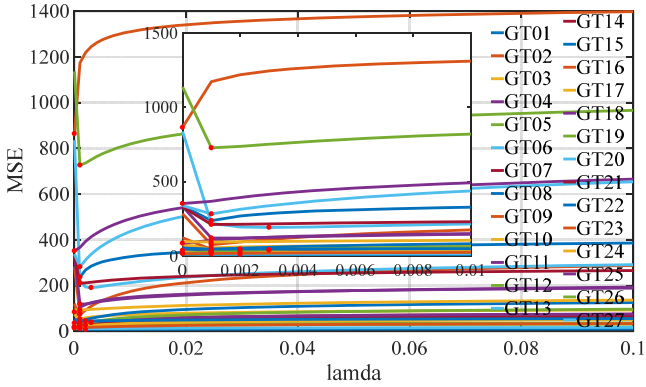

Fig.5 MSE indicator changes with parameter $\lambda$. (a) using trimaps of coarse level 1. (b) using trimaps of coarse level 2.

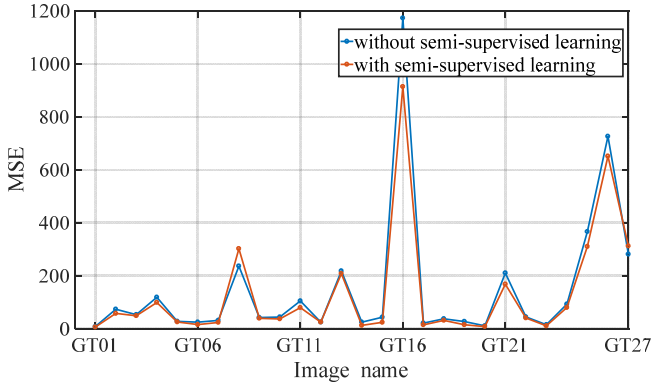

Fig.6.The MSE of proposed matting methos with and without semi-supervised learning.

Fig.7 shows the percentage increase of matting performance(PIMP), on each image by the use of semi-supervised learning iterations. The specific formula for defining *PIMP* is,

$$PIMP = \begin{cases} 1 - \dfrac{MUSL}{MWUSL} & PIMP > 0 \\ 0 & PIMP <= 0 \end{cases}, \qquad (10)$$

where MUSL is the MSE of using semi-supervised learning, MWUSL is the MSE of without using semi-supervised learning. It can be seen that with the increase of iterations, the PIMP

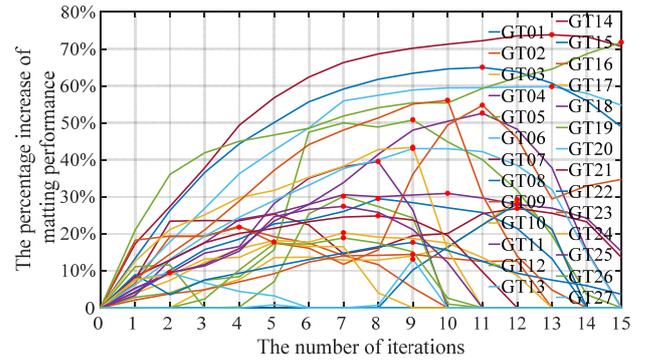

Fig.7. The percentage increase of matting performace(PIMP) changes with the number of iterations

of matting results increase first and then begin to decrease after a certain number of iterations.

In the forepart of iterations, those automatically given labels to originally unknown pixels are correct, so that the matting results are improved by the additional information. When the number of iterations become big, those left unknown pixels are less and tend to be mixed pixels (near the boundary of foreground and background). This makes predicting their correct label very difficult and once some pixels are wrongly labeled, the matting results began to go bad.

This observation enlightens us not using too big iteration numbers and the best iteration number for an image is related not only with the coarseness of its trimap but also some other characteristics (we need to find out in future research).

## IV. CONCLUSION

This paper proposed a matting method based on normalized weight and semi-supervised learning.

The normalized parameter can well control the relative weight between data term and local smooth term in matting. A experimental value range has been suggested for setting this parameter. Semi-supervised learning iterations could significantly reduce users' burden to delineate a refined trimap and get good matting result from a coarse trimap. But the best number of iterations depend not only on the roughness of the trimap, and so is not easy to set. Generally, the more rough the trimap, the more semi-supervised leaning iterations could be taken.

Our future research will focus on adaptively selecting the optimal weight coefficient and the number of semi-supervised learning iterations.

ACKNOWLEDGMENT

This work was partially supported by NSF of China (41161065), NSF of Guizhou (GZKJ [2017]1128) and Educational Commission of Guizhou (KY[2016]027)